\documentclass[letterpaper]{article} 
\usepackage{aaai23}  
\usepackage{times}  
\usepackage{helvet}  
\usepackage{courier}  
\usepackage[hyphens]{url}  
\usepackage{graphicx} 
\usepackage{natbib}  
\usepackage{caption} 
\usepackage{algorithm}
\usepackage{algorithmic}
\usepackage{multirow}
\usepackage{graphicx}
\usepackage{caption}
\usepackage{natbib}
\usepackage{helvet}
\usepackage{amsmath}
\usepackage{amssymb} 
\usepackage{booktabs}
\usepackage{amsfonts}

\usepackage{newfloat}
\usepackage{listings}
\DeclareCaptionStyle{ruled}{labelfont=normalfont,labelsep=colon,strut=off} 
\floatstyle{ruled}
\newfloat{listing}{tb}{lst}{}
\floatname{listing}{Listing}
\title{Spatio-Temporal Graph Neural Point Process for Traffic Congestion Event Prediction}

\author{Guangyin Jin$^{1}$, Lingbo Liu$^{2}$, Fuxian Li$^{3}$, Jincai Huang$^{1}$\\}

\affiliations {
    \textsuperscript{\rm 1} School of System Engineering, National University of Defense Technology \\
    \textsuperscript{\rm 2} Department of Computer Sciences, The Hong Kong Polytechnic University \\
    \textsuperscript{\rm 3} Department of Electronic Engineering, Tsinghua University\\
jinguangyin18@nudt.edu.cn, liulingbo918@gmail.com, fuxianli2020@163.com, huangjincai@nudt.edu.cn 
}

\usepackage{bibentry}

\begin{document}

\maketitle

\begin{abstract}
Traffic congestion event prediction is an important yet challenging task in intelligent transportation systems. Many existing works about traffic prediction integrate various temporal encoders and graph convolution networks (GCNs), called spatio-temporal graph-based neural networks, which focus on predicting dense variables such as flow, speed and demand in time snapshots, but they can hardly forecast the traffic congestion events that are sparsely distributed on the continuous time axis.  In recent years, neural point process (NPP) has emerged as an appropriate framework for event prediction in continuous time scenarios. However, most conventional works about NPP cannot model the complex spatio-temporal dependencies and congestion evolution patterns. To address these limitations, we propose a spatio-temporal graph neural point process framework, named STGNPP for traffic congestion event prediction. Specifically, we first design the spatio-temporal graph learning module to fully capture the long-range spatio-temporal dependencies from the historical traffic state data along with the road network. The extracted spatio-temporal hidden representation and congestion event information are then fed into a continuous gated recurrent unit to model the congestion evolution patterns. In particular, to fully exploit the periodic information, we also improve the intensity function calculation of the point process with a periodic gated mechanism. Finally, our model simultaneously predicts the occurrence time and duration of the next congestion. Extensive experiments on two real-world datasets demonstrate that our method achieves superior performance in comparison to existing state-of-the-art approaches.
\end{abstract}

\section{Introduction}\label{sec:intro}

Traffic congestion is one of the most serious problems in urban management, which is associated with more than 60\% world-wide traffic accidents~\cite{jain2012road}. Since traffic congestion is a continuous process from generation to dissipation, each individual congestion event can be defined by two core elements: occurrence time and duration. Therefore, it is meaningful to predict when the next congestion event will occur and how long it will last for improving the traffic management and scheduling. 

\begin{figure}[htb]
\centering
\includegraphics[width=0.48 \textwidth]{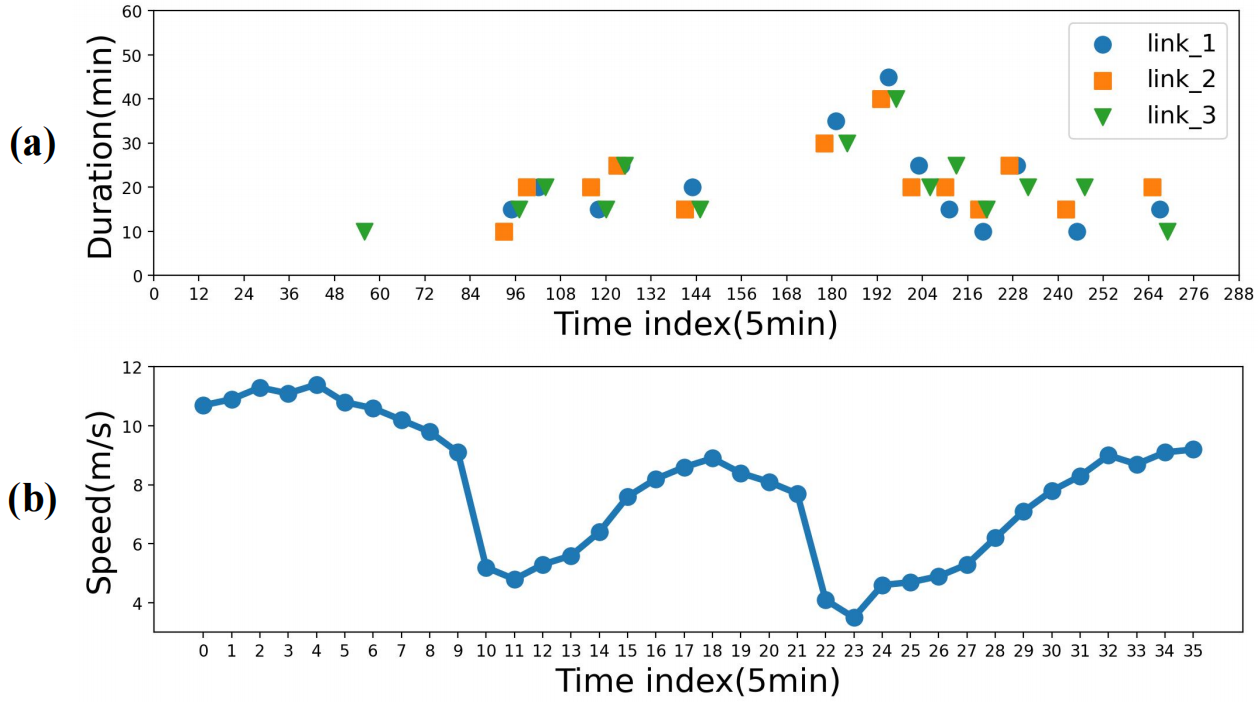}
\caption{Example of the traffic congestion features and link speed trends from the Beijing dataset we adopted in this paper. In sub-figure (a), we select the traffic congestion statistics of three neighbor links on 12 May 2021 to visualize the occurrence time and duration in 24 hours. In sub-figure (b), we select the speed of link 1 from 7.am to 10.am on 12 May 2021 to visualize the change trend.}
\label{fig:intro} 
\end{figure}
In recent years, many works have made some breakthroughs in intelligent transportation, especially the traffic flow prediction~\cite{stgcn,dcrnn,gwn,stgode,stsgcn,gman,stfgnn}. 
Most of these works adopt the architecture of spatio-temporal graph-based neural networks to capture the spatio-temporal correlations, and predict the states of links in future time slots by the homogeneous features in historical regular sequential time slots. Although this framework can fully exploit the spatio-temporal information to improve prediction, it is difficult to handle the traffic congestion event prediction task. There are at least two disadvantages in predicting congestion event: 1)  The traditional traffic prediction framework only model dense variables on the road such as speed, not sparse ones such as congestion events.
2) Most traditional traffic prediction framework~\cite{jin2022automated,stgcn,gwn,jin2020urban,stgode,stsgcn,stfgnn,wang2021pstn,jin2021spatio} can only support the prediction in the given future time window (e.g., the next one hour),
which is difficult to flexibly predict congestion occurring in arbitrary time intervals.

Neural point process is an appropriate framework for sparse event prediction in continuous-time scenarios~\cite{shchur2021neural}. However, this framework is still difficult to be adopted in traffic congestion event prediction directly. There are still two challenges as follows: 
1) \textbf{How to effectively capture the spatio-temporal dependencies in road networks?} The traffic congestion can propagate in the spatial scale over time, thus each link could be affected by the links to which they are adjacent. The periodic information could also bring significant impacts on the occurrence pattern of traffic congestion. For example, during the peak hours (e.g., 7.am$\sim$ 9.am, 5.pm$\sim$ 7.pm), congestion occurs more frequently, but less frequently during other periods. As shown in Fig.~\ref{fig:intro}(a), the congestion frequency is higher during peak hours, and the patterns of occurrence time and duration of traffic congestion on adjacent links are similar. 
However, most previous works about neural point process~\cite{clstm,rmtpp,thtpp,satpp,ftpp,xiao2019learning} 
have not fully captured spatio-temporal information for traffic congestion prediction. 
2) \textbf{How to effectively model the continuous and instantaneous temporal dynamics simultaneously for each road?} The trend of the road states (e.g., speed) is a hybrid mode with continuous and instantaneous changes. When there is no congestion, the road states change gently, but when the congestion occurs, the road conditions could change instantaneously, as shown in Fig.~\ref{fig:intro}(b). This makes the congestion prediction greatly different from some other event forecasting such as StackOverflow, 911 Calls and Electrical Medical Records in previous works~\cite{rmtpp,thtpp,xiao2019learning}, because these events are completely discrete, they do not have the continuously dynamics characteristics similar to road conditions.

To address the problems above, we propose a novel model named Spatio-Temporal Graph Neural Point Process (STGNPP) for traffic congestion event prediction. To be specific, we introduce Transformer and Graph Convolution Network (GCN) to jointly capture the spatio-temporal dependencies from traffic states data. Then we extract the contextual link representations to incorporate with congestion event information for modeling the history of the point process. To encode the hidden evolution patterns of each road, we present a novel continuous Gated Recurrent Unit (GRU) layer with neural flow architecture. In addition, considering the effect of periodic patterns on congestion events, we propose a fully connected network with periodic gated mechanism to calculate the intensity function of the point process. 
Based on the learned intensity function, we compute the likelihood function of traffic congestion events to support the next prediction. 
Our main contributions in this paper are summarized as follows:
\begin{itemize}
\item To the best of our knowledge, it is the first work to propose  spatio-temporal graph neural point process for traffic congestion event prediction. In particular, our model can simultaneously predict when the next traffic congestion will happen and how long it will last.  
\item We take account into the continuous and instantaneous dynamics in road networks, thus propose continuous GRU to model the sequential congestion events. And we also improve the calculation of intensity function by involving the periodic information.
\item We conduct extensive experiments on two real-world traffic datasets. The experimental results demonstrates that our proposed model significantly outperforms than other methods. 
\end{itemize}

\section{Related Works}
\subsection{Traffic Prediction}
In recent years, many deep learning algorithms based on spatio-temporal graph modeling have been widely introduced in traffic prediction. 
Most of them integrated various spatial graph convolution networks (GCNs) and temporal learning modules to extract the complex spatio-temporal dependencies from the structural data. STGCN~\cite{stgcn} is the first work to combine the GCN with 1D CNN for spatio-temporal learning. ASTGCN~\cite{astgcn} involved the attention mechanism based on STGCN. Both DCRNN~\cite{dcrnn} and T-GCN~\cite{tgcn} integrated the gated recurrent unit (GRU) and GCN for traffic prediction. To capture long-range temporal dependencies, both STGNN~\cite{stgnn} and GMAN~\cite{gman} employed self-attention mechanism. 
Graph WaveNet~\cite{gwn}, AGRNN~\cite{agrnn} and DMSTGNN~\cite{han2021dynamic} proposed adaptive graph convolution to enhance the spatial representation from the predefined graphs. 
To capture the continuous spatio-temporal dynamics, STGODE~\cite{stgode} first combined the neural ODE with the normal GCN. However, most previous works about traffic prediction can only predict the sequential data in regular time snapshots and fixed-length time window. This framework is hard to be directly adopted in traffic congestion prediction because congestion may be absent or sparsely distributed over a fixed-length time window. 

\subsection{Neural Point Process}
Neural point process has been widely applied in event forecasting of different domains such as electronic medical records~\cite{enguehard2020neural}, social web~\cite{okawa2019deep,zhang2021learning} and mobility~\cite{wu2021individual,zhu2021traffic}. RMTPP~\cite{rmtpp} first adopted RNN to encode the historical sequential events to obtain the intensity function.  Mei et al.~\cite{clstm} proposed an exponential decay based continuous LSTM to model the event sequences. Zuo et al.~\cite{thtpp} introduced Transformer as the encoder for long-term sequential event learning. To address calculation of integral term for likelihood optimization, Omi et al.~\cite{ftpp} proposed to model the cumulative intensity function. 
However, these previous works only focus on the temporal point process. To consider the spatio-temporal dynamics, a few spatio-temporal neural point process models~\cite{zhu2021imitation,zhou2021neural,chen2020neural} have been proposed in recent years. However, these works can only handle the applications in the spatio-temporal continuous scenarios, which can not be introduced into the traffic congestion event prediction because the traffic networks are discrete in the spatial scale.

\section{Preliminaries}
\subsection{Task Definition}
Given a road network with $N$ links $V(\vert V\vert = N)$, it can be defined as a graph $\mathcal{G} = (V, E, A)$. $E$ denotes the set of edges, whose connections between different links are characterized by the adjacency matrix $A$. The traffic states $\mathcal{X}_{n}$ (eg., link speed) on each link $V_{n}$ are dense features in the snapshots of certain time granularity.
The sequential congestion events on each link $V_{n}$ can be defined as a finite set $\mathcal{S}_{n} = \{s_{n,i}\}(i = 1,2,\dots,|\mathcal{S}_{n}|)$, where $|\mathcal{S}_{n}|$ denotes the length of the set. In this paper, each congestion event can be defined as a two-element tuple $s_{n,i} = \langle t_{n,i}, d_{n,i} \rangle$, where $t_{n,i}$ and $d_{n,i}$ respectively denote the occurrence time and the duration of the $i_{th}$ congestion event on link $V_{n}$. Given a fixed-length historical time window $T$ for each sample, the traffic congestion event prediction task aims to predict the occurrence time and duration of the next congestion event based on the historical congestion events and traffic states. 

\subsection{Point Process Definition}
The point process is one type of stochastic process to simulate the sequential events in a given observation time interval $[0,T]$. The process can be characterized by the conditional intensity function $\lambda(t|H_t)$, which represents the intensity function of events at time point $t$ depended on the historical sequential events $H_t$ up to $t$. The computation of intensity function can be given as:
\begin{equation}
\small
\lambda(t|H_t) = \lim_{\Delta t\to 0}\frac{P(\mbox{one event occurs in } [t,t+\Delta t) | H_t )}{\Delta t}, \label{eq:cif}
\end{equation}
When the conditional intensity function and the time points $\{t_1,t_2,\ldots,t_i\}$ of the historical events are given, the probability density function can be obtained as follows:
\begin{equation}
\small
p(t_{i+1}|t_1,t_2,\ldots,t_i) = \lambda(t_{i+1}|H_{t_{i+1}}) \exp \left\{ -\int_{t_i}^{t_{i+1}} \lambda(t|H_t)dt \right\}, \label{eq:pdf_i}
\end{equation}
where the exponential term in the above equation denotes the probability that no events occur in the time interval $[t_i,t_{i+1})$. 
In addition, we can also obtain the probability density function to observe an event sequence $\{t_i\}_{i=1}^n$, which is defined as follows:
\begin{equation}
\small
p(\{t_i\}_{i=1}^n) = \prod_{i=1}^n \lambda(t_i|H_{t_i}) \exp \left\{- \int_{0}^\tau \lambda(t|H_t) dt\right\}. \label{eq:pdf}
\end{equation}
where $\tau$ is the inter-event time that illustrates the time interval between two different events. Many previous related works~\cite{rmtpp,ftpp,freetpp, satpp} directly use the inter-event time as the basic feature of each event.
Note that, the most crucial part of the point process is the intensity function, which is difficult to characterize in real-world applications. Hence, neural networks can be a fruitful tool to approximate it.\\

\section{Our Model}
\begin{figure}[htb]
\centering
\includegraphics[width=0.47 \textwidth]{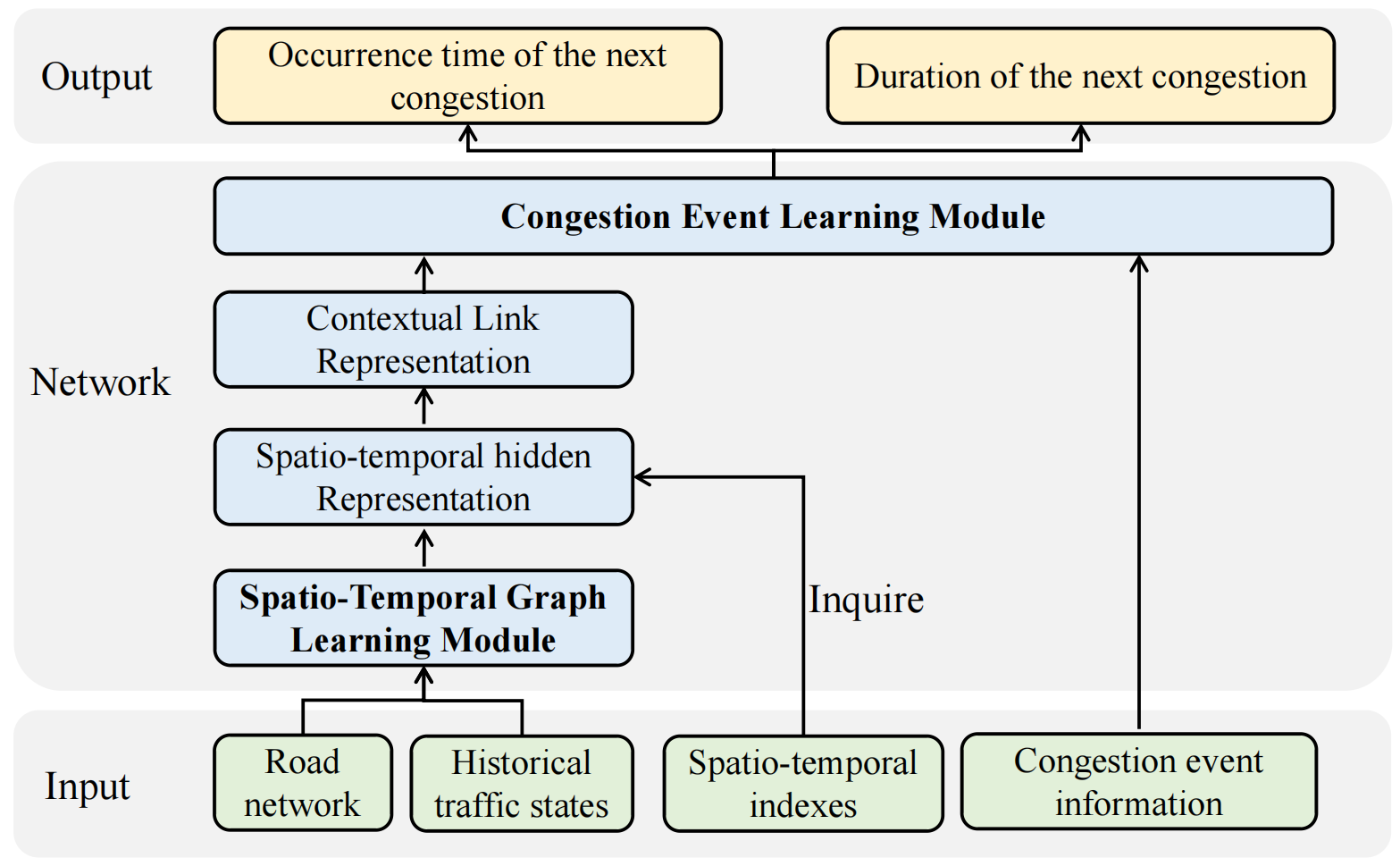}
\caption{The overview of STGNPP}
\label{fig:overview} 
\end{figure}

The overview of our proposed model STGNPP is illustrated in Fig.~\ref{fig:overview}. The initial input data contains four parts: road network, historical traffic states, spatio-temporal indexes and congestion event information. The road network and historical traffic states are fed into spatio-temporal graph learning module to obtain the spatio-temporal hidden representation. The contextual link representation can be extracted from spatio-temporal hidden representation according to the spatio-temporal indexes. And then we integrate the contextual link representation and congestion event information for congestion event learning module. Finally, our model outputs the occurrence time and duration of the next congestion at the same time.

\begin{figure}[h]
\centering
\includegraphics[width=0.47 \textwidth]{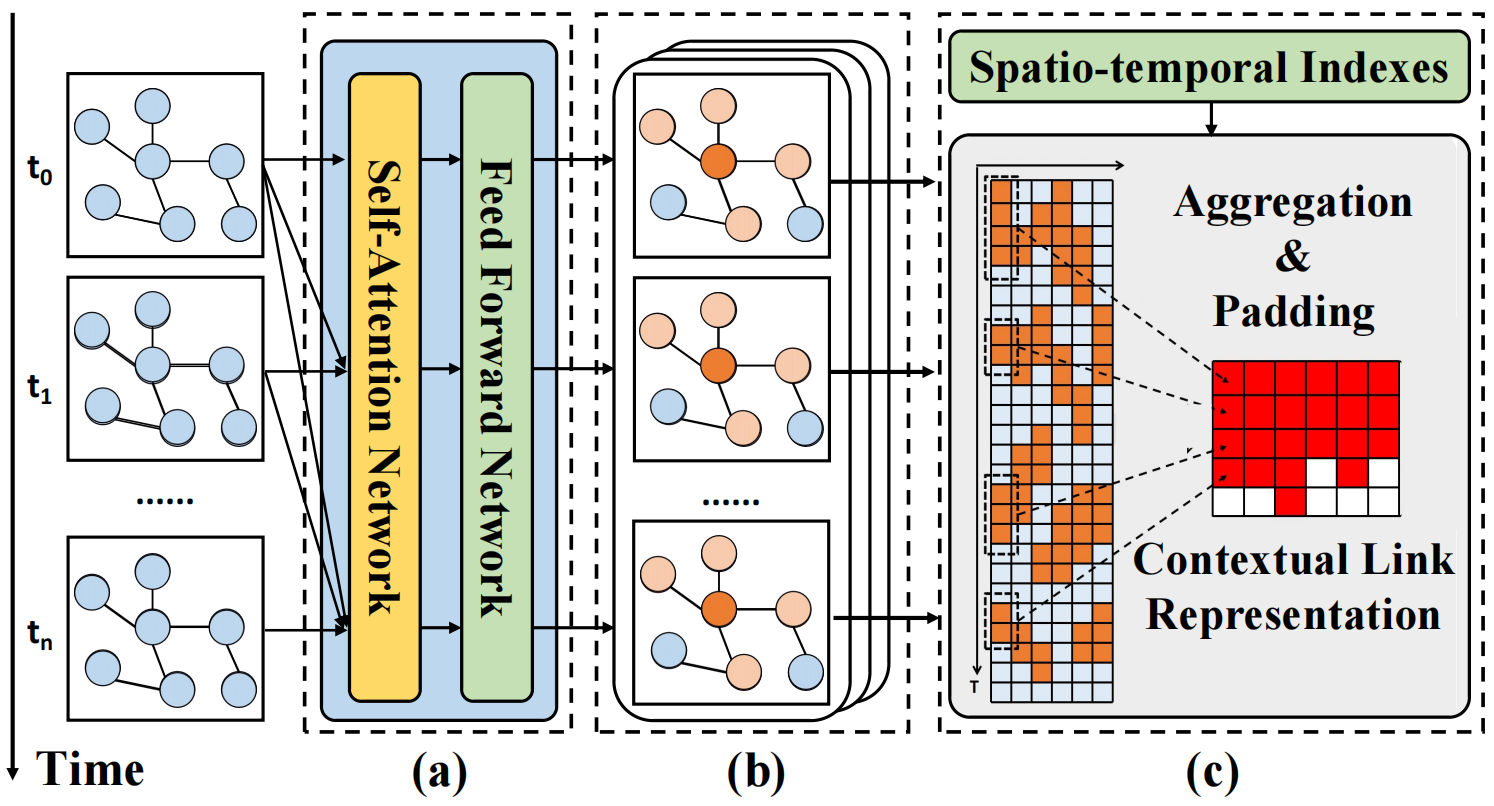}
\caption{The detailed architecture of the spatio-temporal graph learning module. (a) is the link-wise Transformer layer to capture the long-range temporal dependencies.
(b) is the graph convolution layer to further extract the spatial dependencies.
(c) is the spatio-temporal inquirer that can select the corresponding hidden representations according to the spatio-temporal indexes. The spatio-temporal indexes characterize when and where the historical traffic congestion events occupied (time slots painted orange). 
Then we aggregate the latent representations for each congestion event to obtain the contextual link representations.}
\label{fig:st_overview} 
\end{figure}

\subsection{Spatio-Temporal Graph Learning Module}
First, we adopt a fully connected layer to map the historical traffic states into high-dimensional representation $Z\in R^{N\times T\times D}$. $N$, $T$ and $D$ represent the number of links, the length of time window and hidden dimension respectively. The detail architecture of the spatio-temporal graph learning module is illustrated in Fig.~\ref{fig:st_overview}.

\subsubsection{Link-Wise Transformer Layer}
Since we need larger time window to include more historical congestion events, we have to learn the long-range temporal dependencies of the traffic states. Compared with temporal convolution networks~\cite{yu2016multi} and recurrent neural networks~\cite{gru,lstm}, Transformer~\cite{transformer} is a more powerful architecture to capture the long-range dependencies.
The framework of link-wise Transformer layer adopted in our model is illustrated in Figure~\ref{fig:st_overview}(a). 
Note that, the Transformer layer is weight-sharing for different links.
To characterize the sequential relations more explicitly, We employ trigonometric functions-based position encoding method~\cite{transformer} in this case.
The core architecture in the Transformer layer is the self-attention network. We pass the input data into it and compute the attention output by:
\begin{align}
& S = M_{D}(SoftMax(\frac{Q\cdot K^{T}}{\sqrt{D}})\cdot V),\label{eq:e2}\\
& Q = W_{Q}\cdot Z,\; K = W_{K}\cdot Z,\; V = W_{V}\cdot Z,
\label{eq:e3}
\end{align}
where $Q$, $K$ and $V$ respectively denote the query, key, and value matrices obtained by three linear transformations $W_{Q}.W_{K},W_{V}\in R^{D\times D}$, where $D$ denotes the dimension of the self-attention network. $Q\cdot K^{T}\in R^{N\times T\times T}$ denotes the dot product over the $T$ dimension. 
$ M_{D}$ represents the mask operation that sets the value of the upper triangle of the attention matrix to 0. This can prevent information of future time steps from being exploited by past time steps. To stabilize the fitting capability of self-attention network, we also employ the multi-head attention mechanism, similar to~\cite{transformer}.
Then we pass the attention output into the two-layer position-wise feed-forward neural network, generating the sequential hidden representation of each time snapshot:
\begin{align}
& H = W_{F2}\cdot(ReLU(W_{F1}\cdot S + b_{F1})) + b_{F2}, \\
& h(t_{i}) = H(:,i,:),
\label{eq:e5}
\end{align}
where $W_{F1}\in R^{D\times D}$, $W_{F2}\in R^{D\times D}$, $b_{F1}\in R^{D}$ and $b_{F2}\in R^{D}$ are learnable parameters of the two-layer feed-forward neural network. $H\in R^{N\times T\times D}$ is the output of the Transformer layer and $h(t_{i})\in R^{N\times D}$ is the hidden representation at time snapshot $t_{i}$.

\subsubsection{Graph Convolution Layer}
In addition to the temporal dependencies of each link, adjacent links may influence each other, thus we employ the graph convolution layer to capture the spatial dependencies. In this case, we involve the adaptive learnable matrices to characterize the spatial relations that cannot be represented by the predefined adjacency matrix.  And we adopt the simple graph convolution operation~\cite{kipf2017semi} with mix-hop aggregation, which is defined as follows:
\begin{align}
& \hat{A} = A + SoftMax(ReLU(\alpha_{1}\dot\alpha_{2}^{T})),\\
& H_{i} = \sigma(\hat{A}\cdot H_{i-1} \cdot \Theta_{i}),\\
& H_{g} = SumPooling(H_{1},H_{2},\dots,H_{i}),
\label{eq:gcn}  
\end{align}
where $A$ is the normalized predefined adjacency matrix, $\alpha_{1},\alpha_{2}\in R^{N\times D'} (D'<< N)$ are two learnable matrices to adaptively characterize the latent spatial relations through the back propagation process, $\Theta_{i}$ is the learnable weight for each convolution layer, $H_{i}\in R^{N\times T\times d}$ is the output from each layer of GCN. Note that, the initial input of the GCN layer, $H_{0}\in R^{N\times T\times D}$ is the hidden states from Transformer layer and $H_{g}\in R^{N\times T\times D}$ is the output of the sum pooling operation from multi-hop hidden states.

\subsubsection{Spatio-temporal Inquirer}
After obtaining the spatio-temporal graph hidden representations from the Transformer layer and graph convolution layer, in order to characterize the latent features of congestion events, we select the corresponding hidden representations according to the spatio-temporal indexes, as shown in Fig.~\ref{fig:st_overview}(c). 
Each spatio-temporal index contains two elements $v_{n}$ and $t_{s}$. Specifically, $v_{n}$ denotes the index of target link and $t_{s}$ denotes the collection of relative time periods of the congestion event in the historical time window $[0,T]$. From these indexes, we can easily obtain the corresponding hidden representations, named contextual link representations, which is defined as:
\begin{align}
t_{s} &= [t_{i},t_{i+1},\dots,t_{i+L_{s}}], t_{i+L_{s}}<T,\nonumber\\
H_{c} &= \Vert^{N} [PAD(\Vert^{L_{e}} [AGG(H_{g}(v_{n},t_{s},:))])].
\label{eq:select_t}
\end{align}
where $H_{g}(v_{n},t_{s},:)\in R^{L_{s}\times D}$ denotes the latent representations of one congestion event on link $v_{n}$, $L_{s}$ is the number of time slots occupied by one congestion event and $L_{s}$ varies for different congestion events. $AGG(\cdot)$ denotes the aggregation function on the dimension $L_{s}$ and we use the simple sum function in this case. $\Vert^{L_{e}}$ denotes the concentrate operation for the congestion event sequences on each link. Note that, for different links, the lengths of congestion event sequences $L_{e}$ are different, thus we need to adopt the zero padding operation $PAD(\cdot)$ to ensure that the sequences are of equal length $L_{max}$. $\Vert^{N}$ denotes the concentration for the latent representations on different links and the size of output $H_{c}$ is ${N\times L_{max}\times D}$.

\subsection{Congestion Event Learning Module}
From the spatio-temporal learning module, we can obtain the contextual link representation $H_{c}$ for sequential congestion events. Therefore, we define the congestion event representation as follows:
\begin{align}
& H_{e} = W_{e}\cdot [H_{c}, d_{e}] + b_{e},
\label{eq:emb}  
\end{align}
where $d_{e}\in R^{N\times L_{max}\times 1}$ denotes the historical duration of each congestion event after zero padding, 
$H_{e}\in R^{N\times L_{max}\times D}$ is the output representation. 

\begin{figure}[ht]
\centering
\includegraphics[width=0.46 \textwidth]{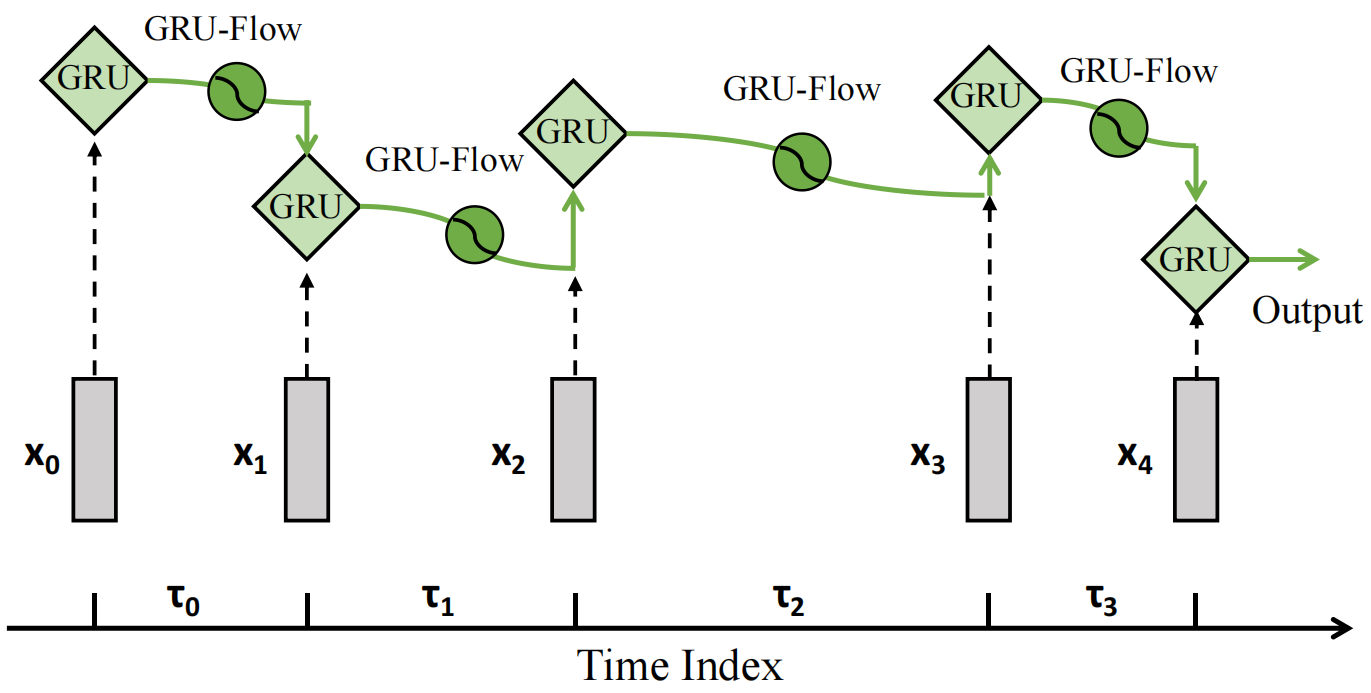}
\caption{The detailed structure of the continuous GRU layer in our model. The green squares denote the moment when the congestion event occurred. The green curves and arrows represent continuous and instantaneous changes in the hidden representation of link states, which are learned by GRU-flow and discrete GRU, respectively. The grey strip denotes the input contextual information at each time step.}
\label{fig:gru_flow} 
\end{figure}

\subsubsection{Continuous GRU Layer}

Different from some other scenarios such as StackOverflow, 911 Calls and Electrical Medical Records in previous related works~\cite{rmtpp,thtpp,xiao2019learning}, the congestion event prediction task is more special: the traffic state for each link is a combination of continuous changes and instantaneous changes. Hence, it is necessary to take these two situations into account. Some continuous system modeling methods are introduced in previous works such as continuous LSTM~\cite{clstm}, neural ODE~\cite{neuralode}, RNN-ODE~\cite{rnn-ode} and GRU-ODE~\cite{GRU-ODE}. However, these traditional methods especially the  ODE-based models suffered from huge computational overhead, which could be a serious problems for real-world applications. Motivated by a recent work~\cite{neuralflow}, we can replace the ODE operations by a more efficient architecture, neural flows, which is defined as:
\begin{equation}
F(t, x) = x + \phi(t)\cdot \Gamma(t,x),
\label{eq:neural_flow}
\end{equation}
where $\phi(t)$ is a continuous function that satisfies two properties: i) $\phi(0)=0$ and ii) $\vert\phi(t)\vert<1$, $\Gamma(t,x)$ is an arbitrary contractive neural network. In this case, we choose \emph{Tanh}, the simplest function for $\phi(t)$. This architecture can transfer the residual connection to the continuous one and reduce the computational overhead brought by ODESolver.
Inspired by GRU-ODE~\cite{GRU-ODE}, the normal GRU Cell can be converted to the continuous type. The equations of normal GRU are defined as:
\begin{align}
  r_t &= \sigma(W_r x_t + U_r h_{t-1}+b_r), \nonumber \\
  z_t &= \sigma(W_z x_t + U_z h_{t-1}+b_z), 
  \label{eq:gru-original} \\
  g_t &= \tanh(W_h  x_t + U_h (r_t \odot h_{t-1}) +  b_h), \nonumber\\
  h_t &= z_t \odot h_{t-1} + (1 - z_t) \odot g_t, \nonumber 
\end{align}
From the above equations, we can derive its differential form:
\begin{align}
	\Delta h_t = h_t - h_{t-1} = (1 - z_t) \odot (g_t - h_{t-1}),
\end{align}
Hence, we propose to combine the differential form of GRU with neural flows, named GRU-flow. We obtain the continuous GRU cell to characterize the hidden state between two adjacent time steps in our model as follows:
\begin{align}
\small
	F(\tau_{i}, h_{i}^{l}) = h_{i}^{l} + \phi (W_{\tau}\cdot\tau_{i})\cdot (1 - z([\tau_{i},h_{i}^{l}]))\nonumber \\ \odot (g([\tau_{i},h_{i}^{l}]) - h_{i}^{l}),
\end{align}
where $\tau_{i}$ is the inter-event time at each step, $W_{\tau}\in R^{1\times D}$ is a transformation weight to ensure dimensional consistency, $h_{i}^{l}$ denotes the $l_{th}$ layer's hidden state at each step, $[\cdot]$ denotes the concentrate operation. The initial input of the continuous GRU cell is the event embedding at step $i$. For capturing the instantaneous dynamics, we directly adopt the discrete GRU cell, which is as follows:
\begin{align}
h_{i+1}^{0} = GRUCell(H_{e}[i+1,:], F(\tau_{i}, h_{i}^{l})),
\end{align}
where $GRUCell(\cdot)$ denotes the discrete GRU cell. The two input elements of discrete GRU cell are respectively the hidden state from the GRU-flow $F(\tau_{i}, h_{i}^{l})$ and the contextual link representation $H_{e}[i+1,:]$ at the next time step. The output of the discrete GRU cell can be treated as the initial input for the GRU-flow at the next time step. The detailed structure of our continuous GRU Layer is show in Fig.~\ref{fig:gru_flow} and the hidden states from the discrete GRU cell are fed into the intensity function network. 

\subsubsection{Intensity Function Network}
To approximate the distribution of inter-event time,  many traditional models~\cite{rmtpp,thtpp} optimized the logarithmic form of the probability density function in eq.~\eqref{eq:pdf}, called log-likelihood function, which is defined as:
\begin{equation}
\log L(\{\tau\} ) = \sum_i \left[ \log \lambda(\tau|h_i) - \int_{0}^{\tau} \lambda(\tau|h_i) d\tau \right]. \label{eq:ll-hazard}
\end{equation}
where $\tau$ denotes the inter-event time, $h_i$ denotes the event hidden state at $i_{th}$ step.
However, the integral term can only be approximated by interpolation or Monte Carlo method~\cite{thtpp}. To address this problem, the cumulative intensity function is introduced to reformulate the log-likelihood function and the intensity function is computed by a multi-layer fully connected network. Since the congestion frequency could vary by peak hours or non-peak hours, as discussed in the introduction section, the intensity function could be impacted by the periodic patterns. To characterize the effect of periodic patterns, we first propose a periodic gated unit to adjust the intensity function. The mathematical form can be defined as the basic intensity term multiplied by the periodic gate term, which is as follows:
\begin{small} 
\begin{align}
\log L(\{t_i\} ) &= \sum_i \left[ \log \left\{ \frac{ \partial}{\partial \tau} \lambda(\tau = t_{i+1}-t_i |h_i) \right\} - \lambda(\tau|h_i) \right], \nonumber \\
\lambda(\tau|h_i) &= f_{l}^{+}([h_i,\tau])\odot \sigma (f_{p}([P_{i}^{d},P_{i}^{w}])).
\label{eq:ll-ch}
\end{align}
\end{small} 
where $f_{l}^{+}(\cdot)$ denotes the fully connected layer for computing the basic intensity function, $f_{p}(\cdot)$ denotes the fully connected layer for periodic gated unit, $P_{i}^{d},P_{i}^{w}$ respectively denote the time of day and the day of week of the $i_{th}$ event. 

\subsection{Optimization and Prediction}
As a multi-task learning framework, our model simultaneously optimizes the negative log-likelihood of the probability density function of the inter-event time and the absolute error of the duration prediction during the training phase:
\begin{equation}
\small
L = \sum_i \left[\lambda(\tau|h_i) - \log \left\{ \frac{ \partial}{\partial \tau} \lambda(\tau|h_i) \right\}\right] + \alpha\Vert f_{d}(h_i)-d_{i+1} \Vert,
\label{eq:loss}
\end{equation}
where $f_{d}(\cdot)$ denotes the fully connected layer for duration prediction of the next traffic congestion, $\alpha$ denotes the trade-off ratio. We set $\alpha$ as 1 by default.

During the inference phase, the duration prediction of the next congestion can be obtained directly by the output of  $f_{d}(\cdot)$. For the occurrence time of the next congestion event, our model can also efficiently generate the prediction in the following way. Given the historical congestion event occurred at time point $\{t_1,t_2,\ldots,t_{i}\}$, the predictive probability density function $p^*(t|t_1,t_2,\ldots,t_{i})$ of the time point $t_{i+1}$ of the next congestion event is calculated from eq.~\eqref{eq:pdf_i}.
In this case, we utilize the median time point $t_{i+1}^*$ of the probability density distribution $p^*$ to estimate $t_{i+1}$. Hence, bisection method is used here and a accurate prediction can be obtained through multiple rounds of iterations

\section{Experiments}

\subsection{Datasets and Settings}
We evaluate our model on Beijing dataset and Chengdu dataset which are collected from the mobile application Amap, as shown in Table~\ref{tab:data}.
Each dataset is chronologically split with 60\% for training, 20\% for validation and 20\% for testing.
We utilize the traffic states, congestion event information and spatio-temporal indexes in the last six hours to predict the occurrence time and duration of the next congestion event. Note that, we only employ the link speed data with condition labels as traffic states data. The condition label is a binary variable that describes whether the road is congested or not for each time slot (5 minutes). And the congestion events are also collected based on the condition labels. Congestion event information includes inter-event times, duration and periodic features.
Our model is implemented by Pytorch 1.5 with NVIDIA TESLA V100 GPU. 
We set the number of stacks of Transformer and GCN as 2, the number of self-attention heads as 4 in Transformer layer, the number of GCN layers as 2 and the number of flow layers in continuous GRU as 2 by default. The dimension of hidden representations in our model is set as 64 and the dimension of random matrix for adaptive graph is set as 10. 
We set the optimizer as Adam with learning rate 0.001 and the batch size as 16. 

\begin{table}[ht]
\footnotesize
\centering
	\scalebox{1}{
	\begin{tabular}{cccc}
		\hline
		Datasets  & \# links & \# Congestion  &Time Range \\ \hline
		Beijing    & 573     & 249464        & 5/12/2021 - 11/12/2021 \\
		Chengdu    & 435     & 204768         & 5/12/2021 - 11/12/2021\\ \hline
	\end{tabular}}
\caption{Dataset description and statistics.}
\label{tab:data}
\end{table}

\begin{table*}[tb]
\footnotesize
\centering
  \label{tab:performance}
  \begin{tabular}{c|ccc|ccc}
	\toprule
			  & \multicolumn{3}{c}{Beijing} & \multicolumn{3}{c}{Chengdu} \tabularnewline
			  Method & NLL  & MAE-t(min)  & MAE-d(min) & NLL   & MAE-t(min)  & MAE-d(min)\tabularnewline
		\midrule
    	HA & /  & 25.45  & 18.70 & /  & 36.35  & 42.67 \tabularnewline
		GBDT & /  & 23.81 $\pm$ 0.56  & 16.21 $\pm$ 0.25 & /  & 32.39 $\pm$ 0.72  & 38.13 $\pm$ 0.63 \tabularnewline
		GRU & /  & 23.18 $\pm$ 0.67 & 15.75 $\pm$ 0.34 & /  & 30.67 $\pm$ 0.69  & 37.45 $\pm$ 0.68 \tabularnewline
		DCRNN & /  & 19.36 $\pm$ 0.52 & 14.63 $\pm$ 0.28 & /  & 27.76 $\pm$ 0.55  & 35.75 $\pm$ 0.53\tabularnewline
		Graph WaveNet & /  & 19.01 $\pm$ 0.16  & 14.46 $\pm$ 0.10 & /  & 27.48 $\pm$ 0.22  & 35.79 $\pm$ 0.16\tabularnewline
		STGODE & /  & 19.68 $\pm$ 0.32  & 14.55 $\pm$ 0.27 & /  & 28.27 $\pm$ 0.34  & 36.69 $\pm$ 0.29\tabularnewline
		NHTPP & 5.84 $\pm$ 0.04  & 19.17 $\pm$ 0.11  & 14.47 $\pm$ 0.11 & 5.90 $\pm$ 0.06  & 28.10 $\pm$ 0.21  & 36.59 $\pm$ 0.15 \tabularnewline
		RMTPP & 6.02 $\pm$ 0.07  & 19.64 $\pm$ 0.13  & 14.52 $\pm$ 0.09 & 6.18 $\pm$ 0.06  & 28.65 $\pm$ 0.23  & 36.45 $\pm$ 0.17 \tabularnewline
		THPP & 5.75 $\pm$ 0.04  & 19.27 $\pm$ 0.14 & 14.38* $\pm$ 0.11 & 5.94 $\pm$ 0.05  & 28.39 $\pm$ 0.20  & 35.86 $\pm$ 0.13 \tabularnewline
		FNN-TPP & 5.46* $\pm$ 0.05  & 18.86* $\pm$ 0.11  & 14.41 $\pm$ 0.08 & 5.68* $\pm$ 0.04  & 27.13* $\pm$ 0.14  & 35.75* $\pm$ 0.06 \tabularnewline
		STGNPP (ours) & \textbf{4.87 $\pm$ 0.03}   & \textbf{16.95 $\pm$ 0.08}  & \textbf{13.15 $\pm$ 0.05}  & \textbf{5.02 $\pm$ 0.02}   & \textbf{24.52 $\pm$ 0.10}  & \textbf{32.94 $\pm$ 0.08} \tabularnewline
		\bottomrule
  \end{tabular}
  \caption{Performance comparison of baseline models and STGNPP. We run all machine learning algorithms five times with different random seeds and calculated the mean and standard deviation. Note that, only our model and neural point process-based models need to compute NLL and the sub-optimal results are marked by the asterisk.}
  \label{tab:performance}
\end{table*}


\begin{table}[ht]
\footnotesize
\centering
	\begin{tabular}{clccc}
		\hline
		Dataset     & Model\&Variants   & NLL     & MAE-t      & MAE-d      \\ \hline
		\multirow{6}{*}{Beijing} & STGNPP        & \textbf{4.87}   & \textbf{16.95}  & \textbf{13.15}         \\
		& GWNPP   & 4.98  & 17.12  & 13.43 \\
		& w/o GCN   & 5.06  & 18.17 & 13.92 \\
		& w/o Trans & 5.17  & 18.45  & 14.10 \\
		& w/o GRU  & 5.43  & 18.68  & 14.32 \\
		& w/o Continous  & 5.01  & 18.38  & 13.94 \\ 
		& w/o Gated  & 5.05  & 18.29  & 13.89 \\ \hline
		\multirow{6}{*}{Chengdu} & STGNPP  & \textbf{5.02}  & \textbf{24.52} & \textbf{32.94}         \\
		& GWNPP   & 5.11  & 24.91  & 33.36 \\
		& w/o GCN   & 5.36  & 26.34  & 35.43 \\
		& w/o Trans & 5.58  & 27.10  & 35.62 \\
		& w/o GRU  & 5.73  & 27.89  & 35.81 \\
		& w/o Continous  & 5.24  & 27.65  & 34.97 \\ 
		& w/o Gated  & 5.18  & 26.11  & 34.95 \\ \hline
	\end{tabular}
 \caption{Ablation experiments.}
	\label{tab:ablation}
\end{table}

\subsection{Overall Performance}
We compare our model with ten state-of-art baselines, which can be divided into three categories. \textbf{Simple models:} Historical Average (HA), Gradient Boosting Decision Tree (GBDT)~\cite{gbdt} and GRU~\cite{gru}. \textbf{Spatio-temporal graph-based models:}  DCRNN~\cite{dcrnn},  Graph WaveNet~\cite{gwn} and STGODE~\cite{stgode}. \textbf{Neural point process-based models:} NHTPP~\cite{clstm}, RMTPP~\cite{rmtpp}, THPP~\cite{thtpp} and FNN-TPP~\cite{ftpp}. For the simple models, we only use the traffic congestion event information as the input features to directly predict the next congestion events.
For the spatio-temporal graph-based models, similar to most traffic flow prediction tasks, we employ the core architectures of spatio-temporal graph networks for spatio-temporal representation learning and predict the link condition labels in the next six hours (the six-hour time window is long enough to guarantee coverage of next congestion events). And then we can obtain the occurrence time and duration of the next traffic congestion event for each link according to the predicted link condition labels in the future time window. 
For neural point process-based models, we use congestion event information for point process modeling to predict the next congestion events. 

The evaluation metrics are mean absolute errors (MAE) and negative log-likelihood (NLL). Note that, we use NLL and MAE to evaluate the prediction of occurrence time. For the prediction of duration, we only use MAE. In subsequent experiments, MAE-t denotes the metrics for occurrence time while MAE-d denotes the metrics for duration.
From the results in Table~\ref{tab:performance}, we can observe that our model STGNPP consistently outperforms the sub-optimal baselines with around 10\% improvements in terms of all metrics on the two datasets, which demonstrates the superiority of our proposed method.
From the experimental results, 
both of spatio-temporal graph-based baselines and neural point process-based baselines are significantly stronger than simple baselines. This is because the simple baselines can neither model inter-event dependencies nor capture spatio-temporal hidden information. The spatio-temporal graph-based baselines can fully exploit spatio-temporal information but neglect to model the congestion events, while neural point process-based baselines are the opposite.
By contrast, our proposed model can not only fully exploit traffic-related spatio-temporal dependencies, but also model the sequence of congestion events, thereby outperforming other methods with a large margin.

\subsection{Ablation Study}
We conduct ablation study on both of the two datasets to evaluate the effectiveness of each crucial module in our model. As shown in Table \ref{tab:ablation}, we compared STGNPP with following ablation variants: 1) \emph{GWNPP}, which replaces the spatio-temporal graph learning module in our model with that in Graph WaveNet 2) \emph{w/o GCN}, which removes all the GCN layers from our models 3) \emph{w/o Trans}, which removes the Transformer layers from our model.   
4) \emph{w/o GRU}, which replaces the continuous GRU with the fully connected networks. 5) \emph{w/o continuous}, which replaces the continuous GRU with the discrete GRU. 6) \emph{w/o Gated}, which removes the periodic gated unit from the intensity function networks.

From Table~\ref{tab:ablation}, we can find that our complete model STGNPP outperforms all the ablation variants. 
Since Graph WaveNet is a recognized excellent model in traffic prediction, the superiority of our model to \emph{GWNPP} suggests that Transformer can  capture long-term dependencies better than temporal convolutions in traffic congestion prediction scenarios.
Our model significantly outperforms \emph{w/o GCN} and \emph{w/o Trans} can demonstrate that capturing either spatial and temporal dependencies of traffic states in road networks can be beneficial for congestion event prediction. 
The variant \emph{w/o GRU} can illustrate that the continuous GRU layer can capture the contextual correlations in historical congestion event sequences for more accurate prediction. 
To further investigate the effectiveness of continuous modeling of GRU flow architectures, we compare STGNPP with \emph{w/o continuous}. 
The results indicate that either the instantaneous dynamics or the continuous dynamics of the contextual link representations have a non-negligible effect on the prediction of traffic congestion event. 
We also compare our model with the \emph{w/o gated} to investigate the effectiveness of periodic gated unit in intensity function network. The comparison results reflect the importance of involving periodic information into intensity function. 

\begin{figure}[h]
\centering
\includegraphics[width=0.47 \textwidth]{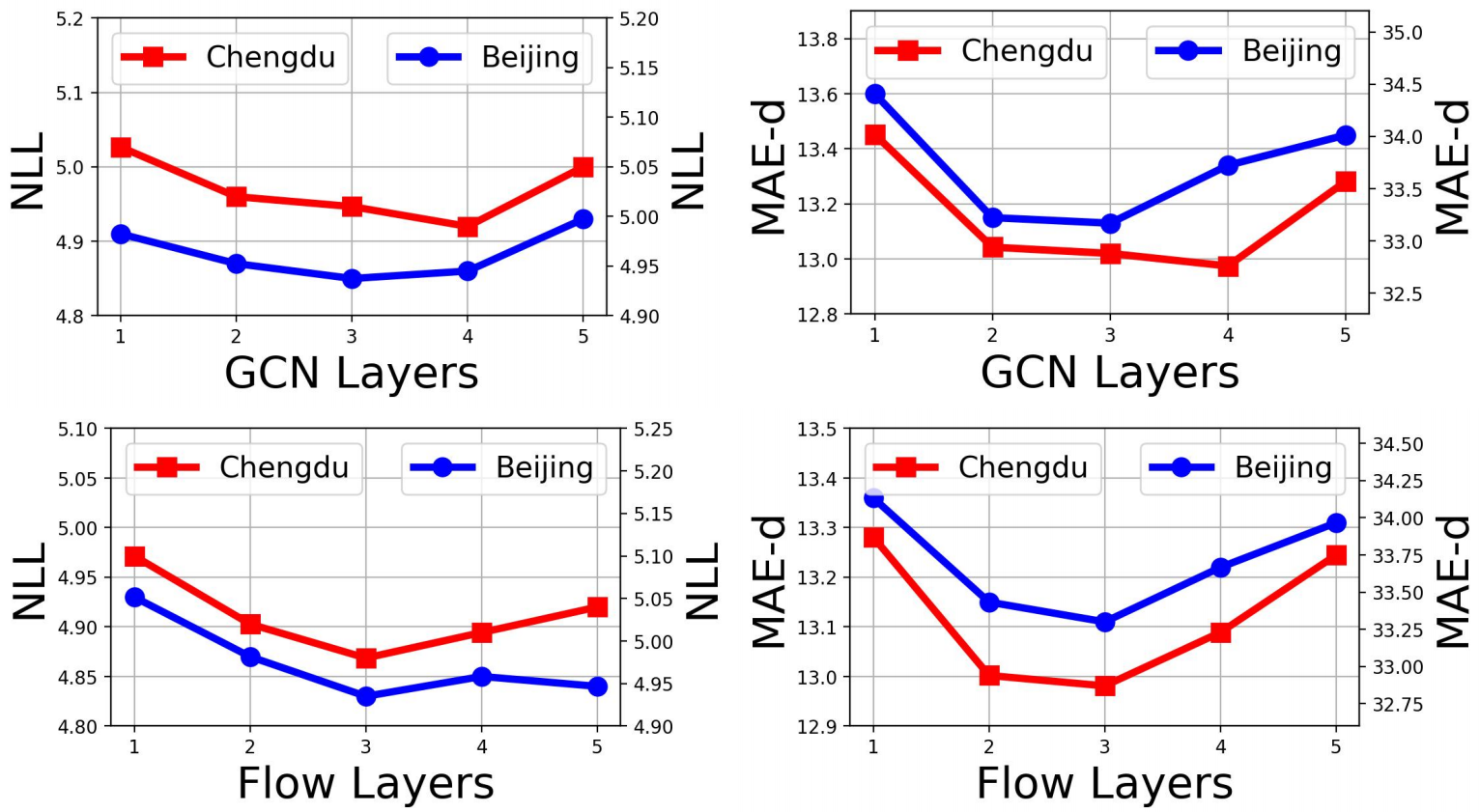}
\caption{Studies on hyper-parameters.}
\label{fig:param} 
\end{figure}

\subsection{Parameters Study}
Since some parameters could significantly affect the learning capability, we conduct parameter study to further investigate the effectiveness of our model. We select the number of GCN layers $l_{g}$ and the number of flow layers $l_{f}$ in continuous GRU, because $l_{g}$ is critical for the spatial dependencies learning and $l_{f}$ is critical for modeling the continuous sequential congestion events. The experimental results are shown in Fig.~\ref{fig:param}. We can find that the best setting for $l_{g}$ is 3 for the two datasets. When $l_{g}$ increases, the NLL of Beijing and duration MAE of the two datasets become worse. The reason may be that the over-smoothing problem of GCNs limits the improvement of performance.
For $l_{f}$, the best value for the two datasets is also 3. With the increase of $l_{f}$,  the NLL of Chengdu and duration MAE of the two datasets become worse. This is because too many layers for continuous GRU could cause the over-fitting problem.


\section{Conclusion}
We propose a novel spatio-temporal graph neural point process framework for traffic congestion event prediction. In this paper, we give a first attempt to utilize the spatio-temporal graph to incorporate with neural point process for traffic congestion event modeling and we also take account into some important traffic-related characteristics such as periodic features, continuous and instantaneous dynamics, to improve the inter-event dependencies learning.
We conduct extensive experiments on two large-scale real-world datasets and the experimental results demonstrate the superiority of our model compared with other traditional methods. 

\bibliography{aaai23}

\begin{thebibliography}{43}
\providecommand{\natexlab}[1]{#1}

\bibitem[{Bai et~al.(2020)Bai, Yao, Li, Wang, and Wang}]{agrnn}
Bai, L.; Yao, L.; Li, C.; Wang, X.; and Wang, C. 2020.
\newblock Adaptive Graph Convolutional Recurrent Network for Traffic
  Forecasting.
\newblock In \emph{NIPS}.

\bibitem[{Bilo{\v{s}} et~al.(2021)Bilo{\v{s}}, Sommer, Rangapuram,
  Januschowski, and G{\"u}nnemann}]{neuralflow}
Bilo{\v{s}}, M.; Sommer, J.; Rangapuram, S.~S.; Januschowski, T.; and
  G{\"u}nnemann, S. 2021.
\newblock Neural Flows: Efficient Alternative to Neural ODEs.
\newblock \emph{NIPS}, 34.

\bibitem[{Brouwer et~al.(2019)Brouwer, Simm, Arany, and Moreau}]{GRU-ODE}
Brouwer, E.~D.; Simm, J.; Arany, A.; and Moreau, Y. 2019.
\newblock GRU-ODE-Bayes: continuous modeling of sporadically-observed time
  series.
\newblock In \emph{NIPS}, 7379--7390.

\bibitem[{Chen, Amos, and Nickel(2020)}]{chen2020neural}
Chen, R.~T.; Amos, B.; and Nickel, M. 2020.
\newblock Neural Spatio-Temporal Point Processes.
\newblock In \emph{ICLR}.

\bibitem[{Chen et~al.(2018)Chen, Rubanova, Bettencourt, and
  Duvenaud}]{neuralode}
Chen, R.~T.; Rubanova, Y.; Bettencourt, J.; and Duvenaud, D. 2018.
\newblock Neural ordinary differential equations.
\newblock In \emph{NIPS}, 6572--6583.

\bibitem[{Cho et~al.(2014)Cho, van Merrienboer, Çaglar G{\"u}lçehre,
  Bahdanau, Bougares, Schwenk, and Bengio}]{gru}
Cho, K.; van Merrienboer, B.; Çaglar G{\"u}lçehre; Bahdanau, D.; Bougares,
  F.; Schwenk, H.; and Bengio, Y. 2014.
\newblock Learning Phrase Representations using RNN Encoder-Decoder for
  Statistical Machine Translation.
\newblock In \emph{EMNLP}.

\bibitem[{Du et~al.(2016)Du, Dai, Trivedi, Upadhyay, Gomez-Rodriguez, and
  Song}]{rmtpp}
Du, N.; Dai, H.; Trivedi, R.; Upadhyay, U.; Gomez-Rodriguez, M.; and Song, L.
  2016.
\newblock Recurrent marked temporal point processes: Embedding event history to
  vector.
\newblock In \emph{Proceedings of the 22nd ACM SIGKDD conference}, 1555--1564.

\bibitem[{Enguehard et~al.(2020)Enguehard, Busbridge, Bozson, Woodcock, and
  Hammerla}]{enguehard2020neural}
Enguehard, J.; Busbridge, D.; Bozson, A.; Woodcock, C.; and Hammerla, N. 2020.
\newblock Neural temporal point processes for modelling electronic health
  records.
\newblock In \emph{Machine Learning for Health}, 85--113. PMLR.

\bibitem[{Fang et~al.(2021)Fang, Long, Song, and Xie}]{stgode}
Fang, Z.; Long, Q.; Song, G.; and Xie, K. 2021.
\newblock Spatial-Temporal Graph ODE Networks for Traffic Flow Forecasting.
\newblock In \emph{Proceedings of the 27th ACM SIGKDD Conference}, 364--373.

\bibitem[{Guo et~al.(2019)Guo, Lin, Feng, Song, and Wan}]{astgcn}
Guo, S.; Lin, Y.; Feng, N.; Song, C.; and Wan, H. 2019.
\newblock Attention Based Spatial-Temporal Graph Convolutional Networks for
  Traffic Flow Forecasting.
\newblock \emph{Proceedings of the AAAI Conference}, 33: 922--929.

\bibitem[{Han et~al.(2021)Han, Du, Sun, Fu, Lv, and Xiong}]{han2021dynamic}
Han, L.; Du, B.; Sun, L.; Fu, Y.; Lv, Y.; and Xiong, H. 2021.
\newblock Dynamic and Multi-faceted Spatio-temporal Deep Learning for Traffic
  Speed Forecasting.
\newblock In \emph{Proceedings of the 27th ACM SIGKDD Conference}, 547--555.

\bibitem[{Hochreiter and Schmidhuber(1997)}]{lstm}
Hochreiter, S.; and Schmidhuber, J. 1997.
\newblock Long Short-term Memory.
\newblock \emph{Neural computation}, 9: 1735--80.

\bibitem[{Jain, Sharma, and Subramanian(2012)}]{jain2012road}
Jain, V.; Sharma, A.; and Subramanian, L. 2012.
\newblock Road traffic congestion in the developing world.
\newblock In \emph{Proceedings of the 2nd ACM Symposium on Computing for
  Development}, 1--10.

\bibitem[{Jin et~al.(2020)Jin, Cui, Zeng, Tang, Feng, and Huang}]{jin2020urban}
Jin, G.; Cui, Y.; Zeng, L.; Tang, H.; Feng, Y.; and Huang, J. 2020.
\newblock Urban ride-hailing demand prediction with multiple spatio-temporal
  information fusion network.
\newblock \emph{Transportation Research Part C: Emerging Technologies}, 117:
  102665.

\bibitem[{Jin et~al.(2022)Jin, Li, Zhang, Wang, and Huang}]{jin2022automated}
Jin, G.; Li, F.; Zhang, J.; Wang, M.; and Huang, J. 2022.
\newblock Automated Dilated Spatio-Temporal Synchronous Graph Modeling for
  Traffic Prediction.
\newblock \emph{IEEE Transactions on Intelligent Transportation Systems}.

\bibitem[{Jin et~al.(2021)Jin, Yan, Li, Huang, and Li}]{jin2021spatio}
Jin, G.; Yan, H.; Li, F.; Huang, J.; and Li, Y. 2021.
\newblock Spatio-Temporal Dual Graph Neural Networks for Travel Time
  Estimation.
\newblock \emph{arXiv preprint arXiv:2105.13591}.

\bibitem[{Kipf and Welling(2017)}]{kipf2017semi}
Kipf, T.~N.; and Welling, M. 2017.
\newblock Semi-supervised classification with graph convolutional networks.
\newblock In \emph{Proc. of ICLR}.

\bibitem[{Li and Zhu(2021)}]{stfgnn}
Li, M.; and Zhu, Z. 2021.
\newblock Spatial-Temporal Fusion Graph Neural Networks for Traffic Flow
  Forecasting.
\newblock \emph{Proceedings of the AAAI Conference on Artificial Intelligence},
  35(5): 4189--4196.

\bibitem[{Li et~al.(2018)Li, Yu, Shahabi, and Liu}]{dcrnn}
Li, Y.; Yu, R.; Shahabi, C.; and Liu, Y. 2018.
\newblock Diffusion convolutional recurrent neural network: Data-driven traffic
  forecasting.
\newblock In \emph{Proc. of ICLR}.

\bibitem[{Mei and Eisner(2017)}]{clstm}
Mei, H.; and Eisner, J.~M. 2017.
\newblock The Neural Hawkes Process: A Neurally Self-Modulating Multivariate
  Point Process.
\newblock \emph{NIPS}, 30.

\bibitem[{Okawa et~al.(2019)Okawa, Iwata, Kurashima, Tanaka, Toda, and
  Ueda}]{okawa2019deep}
Okawa, M.; Iwata, T.; Kurashima, T.; Tanaka, Y.; Toda, H.; and Ueda, N. 2019.
\newblock Deep mixture point processes: Spatio-temporal event prediction with
  rich contextual information.
\newblock In \emph{Proceedings of the 25th ACM SIGKDD Conference}, 373--383.

\bibitem[{Omi, Aihara et~al.(2019)}]{ftpp}
Omi, T.; Aihara, K.; et~al. 2019.
\newblock Fully Neural Network based Model for General Temporal Point
  Processes.
\newblock \emph{NIPS}, 32: 2122--2132.

\bibitem[{Rubanova, Chen, and Duvenaud(2019)}]{rnn-ode}
Rubanova, Y.; Chen, R.~T.; and Duvenaud, D. 2019.
\newblock Latent ODEs for irregularly-sampled time series.
\newblock In \emph{NIPS}, 5320--5330.

\bibitem[{Shchur, Bilo{\v{s}}, and G{\"u}nnemann(2019)}]{freetpp}
Shchur, O.; Bilo{\v{s}}, M.; and G{\"u}nnemann, S. 2019.
\newblock Intensity-Free Learning of Temporal Point Processes.
\newblock In \emph{ICLR}.

\bibitem[{Shchur et~al.(2021)Shchur, T{\"u}rkmen, Januschowski, and
  G{\"u}nnemann}]{shchur2021neural}
Shchur, O.; T{\"u}rkmen, A.~C.; Januschowski, T.; and G{\"u}nnemann, S. 2021.
\newblock Neural Temporal Point Processes: A Review.
\newblock \emph{arXiv preprint arXiv:2104.03528}.

\bibitem[{Song et~al.(2020)Song, Lin, Guo, and Wan}]{stsgcn}
Song, C.; Lin, Y.; Guo, S.; and Wan, H. 2020.
\newblock Spatial-temporal synchronous graph convolutional networks: A new
  framework for spatial-temporal network data forecasting.
\newblock In \emph{Proceedings of the AAAI Conference}, volume~34, 914--921.

\bibitem[{Vaswani et~al.(2017)Vaswani, Shazeer, Parmar, Uszkoreit, Jones,
  Gomez, Kaiser, and Polosukhin}]{transformer}
Vaswani, A.; Shazeer, N.; Parmar, N.; Uszkoreit, J.; Jones, L.; Gomez, A.~N.;
  Kaiser, L.; and Polosukhin, I. 2017.
\newblock Attention Is All You Need.
\newblock \emph{CoRR}, abs/1706.03762.

\bibitem[{Wang, Zhang, and Tsui(2021)}]{wang2021pstn}
Wang, T.; Zhang, Z.; and Tsui, K.-L. 2021.
\newblock PSTN: Periodic Spatial-temporal Deep Neural Network for Traffic
  Condition Prediction.
\newblock \emph{arXiv preprint arXiv:2108.02424}.

\bibitem[{Wang et~al.(2020)Wang, Ma, Wang, Jin, Wang, Tang, Jia, and
  Yu}]{stgnn}
Wang, X.; Ma, Y.; Wang, Y.; Jin, W.; Wang, X.; Tang, J.; Jia, C.; and Yu, J.
  2020.
\newblock Traffic Flow Prediction via Spatial Temporal Graph Neural Network.
\newblock In \emph{Proceedings of The Web Conference 2020}, 1082–1092.
\newblock ISBN 9781450370233.

\bibitem[{Wu, Cheng, and Sun(2021)}]{wu2021individual}
Wu, Y.; Cheng, Z.; and Sun, L. 2021.
\newblock Individual Mobility Prediction via Attentive Marked Temporal Point
  Processes.
\newblock \emph{arXiv preprint arXiv:2109.02715}.

\bibitem[{Wu et~al.(2019)Wu, Pan, Long, Jiang, and Zhang}]{gwn}
Wu, Z.; Pan, S.; Long, G.; Jiang, J.; and Zhang, C. 2019.
\newblock Graph WaveNet for Deep Spatial-Temporal Graph Modeling.
\newblock In \emph{Proc. of IJCAI}.

\bibitem[{Xiao et~al.(2019)Xiao, Yan, Farajtabar, Song, Yang, and
  Zha}]{xiao2019learning}
Xiao, S.; Yan, J.; Farajtabar, M.; Song, L.; Yang, X.; and Zha, H. 2019.
\newblock Learning time series associated event sequences with recurrent point
  process networks.
\newblock \emph{IEEE transactions on neural networks and learning systems},
  30(10): 3124--3136.

\bibitem[{Ye et~al.(2009)Ye, Chow, Chen, and Zheng}]{gbdt}
Ye, J.; Chow, J.-H.; Chen, J.; and Zheng, Z. 2009.
\newblock Stochastic gradient boosted distributed decision trees.
\newblock In \emph{Proceedings of the 18th ACM CIKM Conference}, 2061--2064.

\bibitem[{Yu, Yin, and Zhu(2018)}]{stgcn}
Yu, B.; Yin, H.; and Zhu, Z. 2018.
\newblock Spatio-Temporal Graph Convolutional Networks: A Deep Learning
  Framework for Traffic Forecasting.
\newblock 3634--3640.

\bibitem[{Yu and Koltun(2016)}]{yu2016multi}
Yu, F.; and Koltun, V. 2016.
\newblock Multi-scale context aggregation by dilated convolutions.
\newblock In \emph{ICLR}.

\bibitem[{Zhang et~al.(2020)Zhang, Lipani, Kirnap, and Yilmaz}]{satpp}
Zhang, Q.; Lipani, A.; Kirnap, O.; and Yilmaz, E. 2020.
\newblock Self-attentive hawkes process.
\newblock In \emph{ICML}, 11183--11193. PMLR.

\bibitem[{Zhang, Lipani, and Yilmaz(2021)}]{zhang2021learning}
Zhang, Q.; Lipani, A.; and Yilmaz, E. 2021.
\newblock Learning Neural Point Processes with Latent Graphs.
\newblock In \emph{Proceedings of the Web Conference 2021}, 1495--1505.

\bibitem[{Zhao et~al.(2019)Zhao, Song, Zhang, Liu, Wang, Lin, Deng, and
  Li}]{tgcn}
Zhao, L.; Song, Y.; Zhang, C.; Liu, Y.; Wang, P.; Lin, T.; Deng, M.; and Li, H.
  2019.
\newblock T-gcn: A temporal graph convolutional network for traffic prediction.
\newblock \emph{IEEE Transactions on Intelligent Transportation Systems},
  21(9): 3848--3858.

\bibitem[{Zheng et~al.(2020)Zheng, Fan, Wang, and Qi}]{gman}
Zheng, C.; Fan, X.; Wang, C.; and Qi, J. 2020.
\newblock GMAN: A Graph Multi-Attention Network for Traffic Prediction.
\newblock \emph{Proceedings of the AAAI Conference}, 34: 1234--1241.

\bibitem[{Zhou et~al.(2021)Zhou, Yang, Rossi, Zhao, and Yu}]{zhou2021neural}
Zhou, Z.; Yang, X.; Rossi, R.; Zhao, H.; and Yu, R. 2021.
\newblock Neural Point Process for Learning Spatiotemporal Event Dynamics.
\newblock \emph{arXiv preprint arXiv:2112.06351}.

\bibitem[{Zhu et~al.(2021{\natexlab{a}})Zhu, Ding, Zhang, Van~Hentenryck, and
  Xie}]{zhu2021traffic}
Zhu, S.; Ding, R.; Zhang, M.; Van~Hentenryck, P.; and Xie, Y.
  2021{\natexlab{a}}.
\newblock Spatio-temporal point processes with attention for traffic congestion
  event modeling.
\newblock \emph{IEEE Transactions on Intelligent Transportation Systems}.

\bibitem[{Zhu et~al.(2021{\natexlab{b}})Zhu, Li, Peng, and
  Xie}]{zhu2021imitation}
Zhu, S.; Li, S.; Peng, Z.; and Xie, Y. 2021{\natexlab{b}}.
\newblock Imitation Learning of Neural Spatio-Temporal Point Processes.
\newblock \emph{IEEE Transactions on Knowledge and Data Engineering}.

\bibitem[{Zuo et~al.(2020)Zuo, Jiang, Li, Zhao, and Zha}]{thtpp}
Zuo, S.; Jiang, H.; Li, Z.; Zhao, T.; and Zha, H. 2020.
\newblock Transformer hawkes process.
\newblock In \emph{ICML}, 11692--11702. PMLR.

\end{thebibliography}

\end{document}